\pdfoutput=1

\documentclass[11pt]{article}

 \usepackage{acl}

\usepackage{times}
\usepackage{latexsym}

\usepackage[T1]{fontenc}

\usepackage[utf8]{inputenc}

\usepackage{microtype}

\usepackage{graphicx}
\usepackage{multirow}
\usepackage{lipsum}
\usepackage{array}
\usepackage{booktabs}
\usepackage{float}
\usepackage{hyperref}
\usepackage{xcolor}
\usepackage[normalem]{ulem}
\useunder{\uline}{\ul}{}
\usepackage{amsmath}
\usepackage{amsfonts}
\usepackage{array}

\title{\includegraphics[scale=0.02]{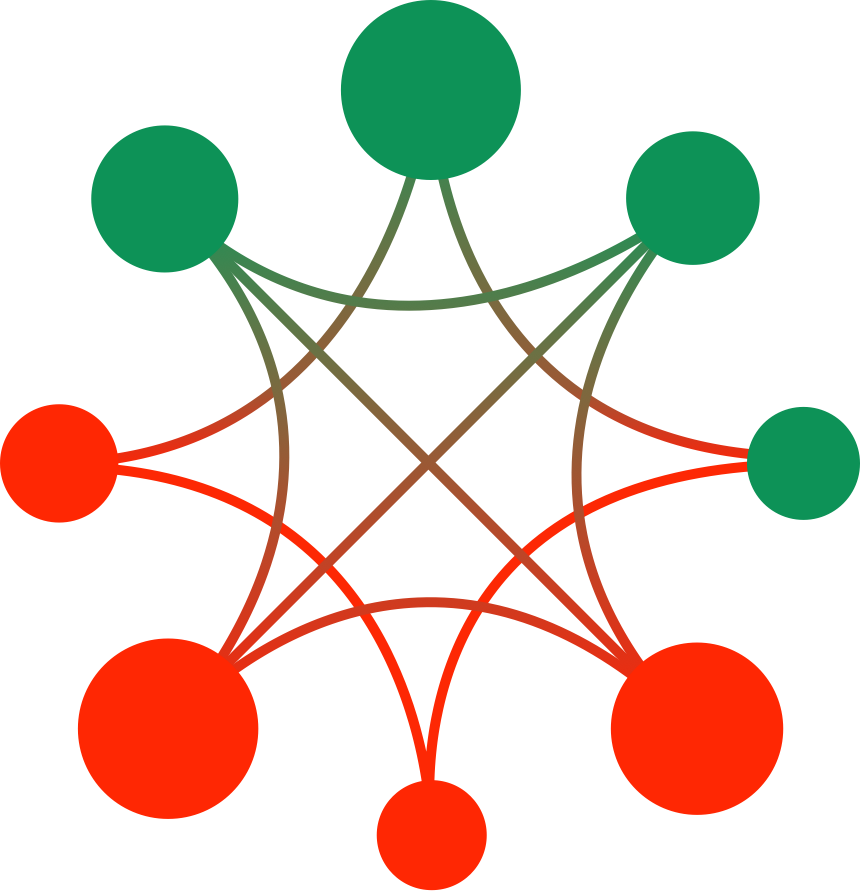} Controlled Text Generation for Large Language Model with Dynamic Attribute Graphs}

\author{\textbf{Xun Liang}$^{1,}$\thanks{Equal contribution}~~~~\textbf{Hanyu Wang}$^{1,*}$~~~~\textbf{Shichao Song}$^{1,*}$ \\
\textbf{Mengting Hu}$^{2}$~~~~\textbf{Xunzhi Wang}$^{2}$~~~~\textbf{Zhiyu Li}$^{3,}$\thanks{Corresponding author}~~~~\textbf{Feiyu Xiong}$^{3}$~~~~\textbf{Bo Tang}$^{3}$ \\
    $^1$Renmin University of China \quad
    $^2$Nankai University \\
    $^3$Institute for Advanced Algorithms Research (Shanghai) \\
    \texttt{\{xliang,hy.wang,songshichao\}@ruc.edu.cn}\\
    \texttt{\{mthu,xunzhi\}@mail.nankai.edu.cn},
    \texttt{\{lizy,xiongfy,tangb\}@iaar.ac.cn}\\
}

\begin{document}
\maketitle
\begin{abstract}

Controlled Text Generation (CTG) aims to produce texts that exhibit specific desired attributes. In this study, we introduce a pluggable CTG framework for Large Language Models (LLMs) named \textbf{D}ynamic \textbf{At}tribute \textbf{G}raphs-based controlled text generation (\textbf{DATG})\footnote{Our code is available at \url{https://github.com/IAAR-Shanghai/DATG}}. This framework utilizes an attribute scorer to evaluate the attributes of sentences generated by LLMs and constructs dynamic attribute graphs. DATG modulates the occurrence of key attribute words and key anti-attribute words, achieving effective attribute control without compromising the original capabilities of the model. We conduct experiments across four datasets in two tasks: toxicity mitigation and sentiment transformation, employing five LLMs as foundational models. Our findings highlight a remarkable enhancement in control accuracy, achieving a peak improvement of 19.29\% over baseline methods in the most favorable task across four datasets. Additionally, we observe a significant decrease in perplexity, markedly improving text fluency.
\\

\textcolor{red}{CONTENT WARNING: This document, for the purpose of illustrating tasks related to toxicity in CTG, may contain examples that are offensive. Please read selectively.}

\end{abstract}

\section{Introduction}

Controlled Text Generation (CTG) focuses on generating text adhering to specific conditions or attributes, such as sentiment, non-toxicity\cite{liu_dexperts_2021, pei_preadd_2023} and style\cite{konen_stylevectors_2024, tao_cat_2024}.
In the realm of CTG, achieving precise control over specific attributes of the generated content is a significant challenge. This must be accomplished without compromising the generative capabilities and text quality of LLMs.

\begin{figure}[t]
  \centering
  \includegraphics[width=0.5\textwidth]{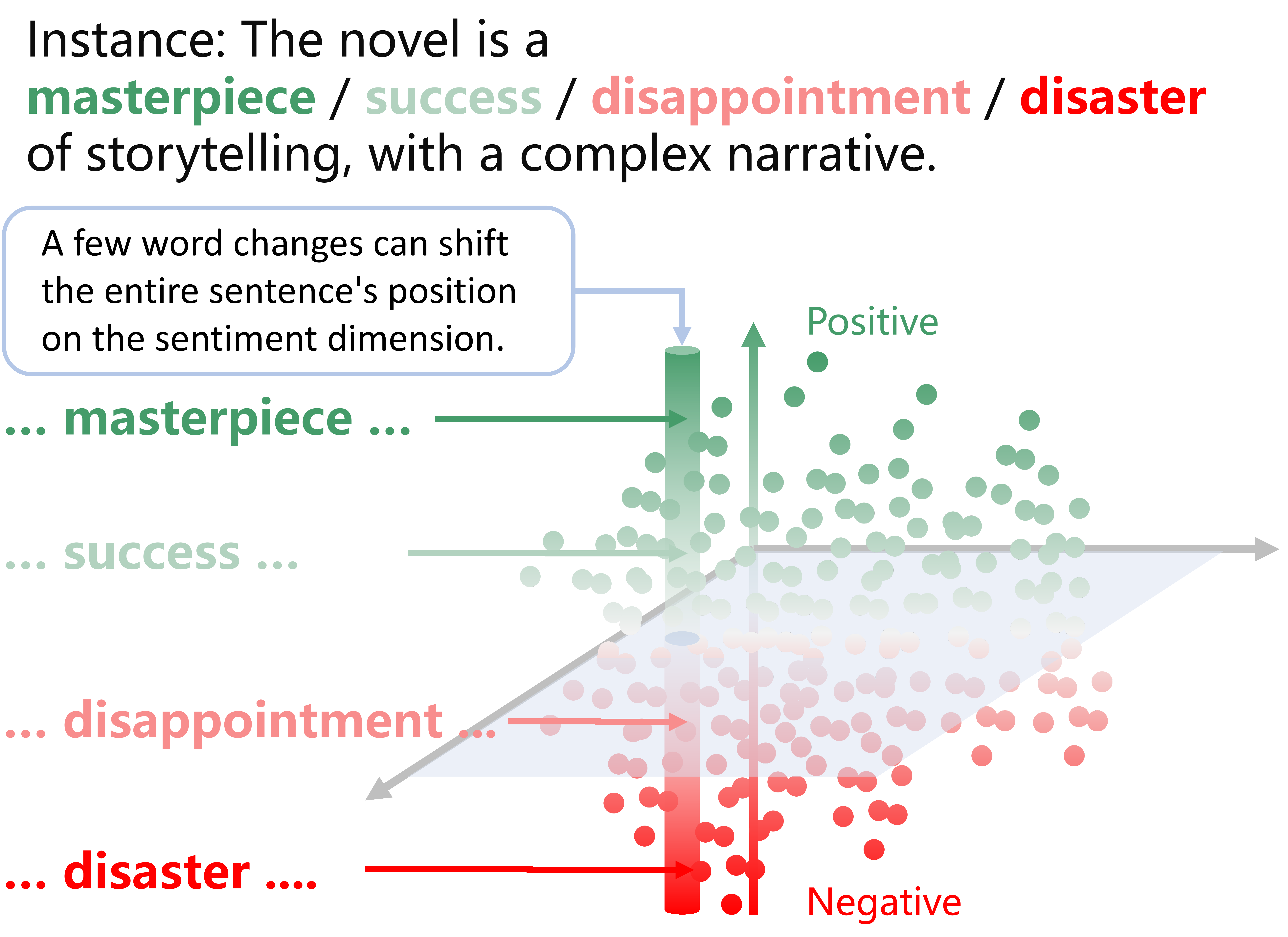}
  \caption{Illustration of the impact of key words on text attributes within the semantic space.}
  \label{fig:semantic_space} 
\end{figure}

Traditionally, CTG methods have employed small language models to influence the decoding process of larger models\cite{dathathri_pplm_2020, krause_gedi_2021, yang_fudge_2021}. Though this approach provides a degree of control, it may compromise the inherent quality and variability of the output.
Recent studies \cite{zhong_air-decoding_2023} highlight that an overemphasis on control can detrimentally affect text fluency, rendering the content less effective. This issue underscores a critical insight: excessive reliance on smaller language models to steer the outputs of LLMs can diminish the decoding capabilities inherent to LLMs. When small-scale models assume control, they effectively overshadow the original performance of LLMs during the inference and decoding phase. This process not only masks the vast capabilities of LLMs but also relegates them to a subordinate role, essentially transforming these sophisticated generative models into mere ``\textit{puppets}'' of their smaller counterparts.

\begin{figure*}[!ht]
    \centering
    \includegraphics[width=1.0\linewidth]{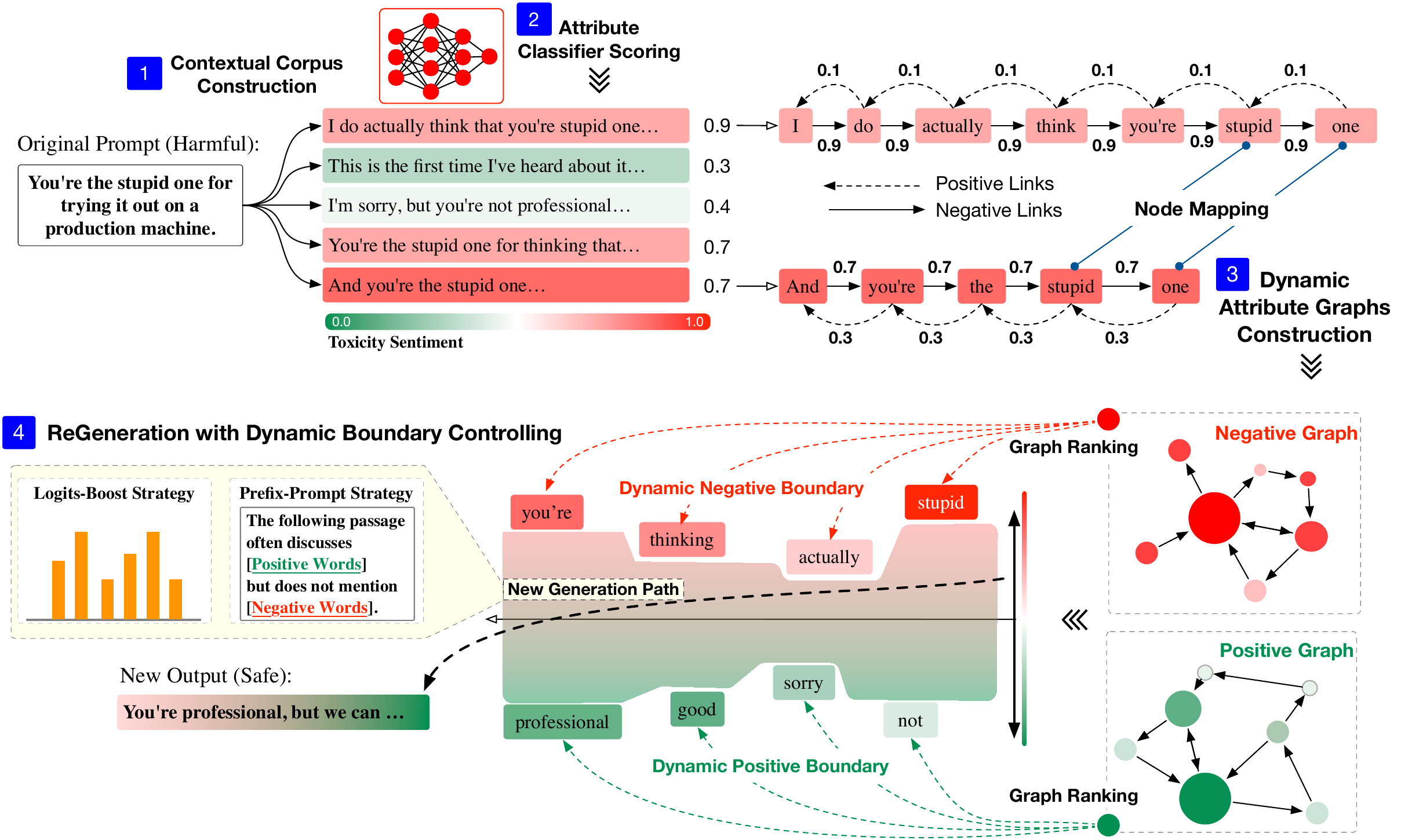}
    \caption{DATG unfolds in four stages: (1) \textbf{Contextual Corpus Construction}, using LLMs to generate text sequences from specified prompts; (2) \textbf{Attribute Classifier Scoring}, employing classifiers to evaluate texts against target attributes; (3) \textbf{Dynamic Attribute Graphs Construction}, forming attribute graphs based on classifier-informed token linkages, encapsulating texts' compliance and divergence from the target attribute in semantic space; (4) \textbf{ReGeneration with Dynamic Boundary Controlling}, applying graph ranking to identify and adjust key nodes, guiding text toward the desired attribute boundary via logits-boost and prefix-prompt strategies.}
    \label{fig:framework}
\end{figure*}

In light of our exploration, we think the specific attributes of a text are predominantly determined by a limited number of words that bear close relation to those attributes \cite{zhong_air-decoding_2023}. Despite these key words being sparse within the text, their impact on the overall attributes is decisive. For instance, changing the word ``\textit{masterpiece}'' to ``\textit{failure}'' in the sentence ``\textit{The novel is a masterpiece of storytelling, with a complex narrative.}'' shifts the sentiment from positive to negative. This change alters the entire sentence's sentiment and meaning. In the conceptual framework of semantic space, these attributes can be seen as dimensions within this space. By strategically adjusting these key words, we can guide the text generated by LLMs to move in the desired direction within the semantic space, controlling its attributes without significant alterations to the overall content (See Figure \ref{fig:semantic_space}).

Based on these observations, we propose a pluggable CTG approach, \textbf{D}ynamic \textbf{At}tribute \textbf{G}raphs-based controlled text generation (\textbf{DATG}), which employs dynamic attribute graphs to identify key words aligned or opposed to target attribute dimensions. By modulating the occurrence of these key words, our method precisely controls text attributes without compromising the inherent capabilities of LLMs. This strategy allows for targeted movement within the semantic space.

As described in Figure \ref{fig:framework}. Our work begins with \textbf{Contextual Corpus Construction}, where LLMs generate text sequences from specific prompts. Subsequently, \textbf{Attribute Classifier Scoring} assesses these texts with classifiers, such as toxicity or sentiment classifiers, to evaluate alignment with the target attribute. 
The core of our method, \textbf{Dynamic Attribute Graphs Construction}, transforms the text sequences into directed weighted graphs, informed by classifier scores. This process leads to the creation of two distinct graphs: a positive attribute graph, weighted by the consistency scores from the classifier, and a negative attribute graph, weighted by the complements of these scores. The attribute graphs represent the text's adherence to and deviation from the target attribute dimension within the semantic space. 
During the \textbf{ReGeneration with Dynamic Boundary Controlling} process, the graph ranking algorithm selects key nodes that propel the generated text towards the upper boundary of the control attribute dimension in the semantic space. Adjustments of the occurrence of these key nodes, facilitated by logits-boost and prefix-prompt strategies, enable the regeneration of text.

The key contributions of our study are summarized as follows:

\begin{itemize}
    \item We introduce a pluggable DATG framework that integrates dynamic attribute graphs with LLMs for CTG, providing a novel, flexible approach to attribute-driven text generation. 
    \item DATG achieves a peak improvement of 19.29\% in performance over baseline methods, according to comprehensive experiments across various datasets, and significantly enhances text fluency. 
    \item We reintroduce the application of graph models in CTG tasks, offering new insights for controlled text generation with LLMs.
\end{itemize}

\section{Methodology}
\subsection{Problem Definition}

The generative capability of LLMs is characterized by the probability distribution over a sequence \(X\):
\begin{equation}
P(x_n | X_{1:n-1}) = p(x_n | x_1, x_2, \ldots, x_{n-1}),
\end{equation}
where \(x_n\) represents the token currently being generated, and \(X_{1:n-1}\) includes the sequence of tokens generated prior to \(x_n\). This probabilistic framework allows LLMs to produce text sequences that are diverse and coherent.

In the domain of CTG, control conditions \(C\) are integrated into the generative process to steer the text towards exhibiting specific attributes, such as sentiment and toxicity. This can be formulated as:
\begin{equation}
P(X | C) = \prod_{i=1}^{n}{p(x_i | x_{<i}, C)},
\end{equation}
where \(C\) signifies the desired attributes to be reflected in the generated text. The key challenge in CTG is to integrate \(C\) into the generative process seamlessly, maintaining the LLMs’ inherent generative quality.

We consider the problem within the framework of a semantic space \(\mathcal{S} \subset \mathbb{R}^d\), where outputs of LLMs are mapped as vectors. In this semantic space \(\mathcal{S}\), our goal is to adjust dimensions associated with control conditions \(C\), directing the distribution of text towards desired attributes while preserving the integrity of other semantic dimensions. This objective is achieved through a transformation function \(f\), designed to delicately shift semantic vectors without altering their inherent characteristics:
\begin{equation}
J(f) = \mathbb{E}_{\mathbf{x} \sim P(\mathcal{S})} [s(f(\mathbf{x}))],
\end{equation}
where \(J(f)\) evaluates the effectiveness of \(f\) in aligning text generation with control conditions \(C\), and \(s(\cdot)\) measures the semantic vector's conformity to these conditions. To depict the vector transition within \(\mathcal{S}\) towards desired attributes, we employ the transformation equation:
\begin{equation}
\mathbf{x}_{\text{after}} = f(\mathbf{x}_{\text{before}}) = \mathbf{x}_{\text{before}} + \Delta\mathbf{x},
\end{equation}

Leveraging attribute graphs, we identify key words that significantly influence the LLM-generated sentences in the semantic space \(\mathcal{S}\), along the control attribute dimension. By adjusting the occurrence of just a few key words, we not only preserve the original performance of LLMs but also effectively steer the regenerated text towards desired conditions. This method effectively guides the text towards specified attributes, maintaining semantic integrity and coherence.

\subsection{Contextual Corpus Construction}

Recent studies, including LIMA \cite{zhou_lima_2023} and Re-Align \cite{lin_realign_2023}, affirm that the foundational knowledge and capabilities of LLMs are established predominantly during the pre-training phase. This evidence suggests that unaligned base models already possess the capacity to generate the desired texts.

Guided by the principles of the LIMA hypothesis and findings from Re-Align, our approach commences with the generation of a sentence set, symbolized as \( \mathbf{X} \), using an LLM prompted by a query that is intricately tied to the desired context. This initial phase leverages the LLM's pre-trained knowledge to generate text sequences closely aligned with the prompt's context, reflecting the inherent distribution of text in the semantic space produced by large language models.

The set comprises individual sentences, \( X_j \), each generated in response to the initial prompt, represented as \( \mathbf{X} = \{X_1, X_2, \ldots, X_m\} \). Each sentence \( X_j \) is a sequence of tokens \( \{x_{1j}, x_{2j}, \ldots, x_{n_jj}\} \), where \( n_j \) denotes the sentence's token count. This constructs a contextual corpus foundational for subsequent manipulations.

\subsection{Attribute Classifier Scoring}

To align generated texts with specific attributes like toxicity or sentiment levels, we employ a pre-trained language model enhanced with a classification layer. This classifier is fine-tuned on data tailored to the target attribute, enabling a condition-specific classifier to precisely evaluate and quantify attribute presence and intensity.

The classifier model scores each text \(X_i\) in \(\mathbf{X} = \{X_1, X_2, \ldots, X_m\}\) as:
\begin{equation}
s(X_i) = \text{ClassifierModel}(X_i),
\end{equation}
where \(s(X_i)\), between 0 and 1, reflects how well \(X_i\) exhibits the target attribute and assesses text distribution along the control condition in the semantic space. This scoring, a quantitative metric, aids in evaluating attribute representation in \(\mathbf{X}\) and understanding text alignment with control conditions.

\subsection{Dynamic Attribute Graphs Construction}

In the dynamic attribute graphs construction phase, each sentence \(X_j\) in \(\mathbf{X}\) is tokenized into discrete tokens, forming vertex sets \(V_j = \{v_{1,j}, v_{2,j}, \ldots, v_{n_j,j}\}\) for each sentence:
\begin{equation}
V = \bigcup_{j=1}^{m} V_j,
\end{equation}
where \(v_{i,j}\) represents a distinct token from sentence \(X_j\), and \(V\) is the union of all vertex sets \(V_j\).

Directed edges within each \(V_j\) are defined by sequentially linking tokens to reflect their order in the sentence:
\begin{equation}
E_j = \{(v_{i,j}, v_{i+1,j}) \mid v_{i,j}, v_{i+1,j} \in V_j\},
\end{equation}
The overall edge set \(E\) is then defined as the union of all \(E_j\), reflecting the aggregation of directed edges from all sentences:
\begin{equation}
E = \bigcup_{j=1}^{m} E_j,
\end{equation}

In the dynamic attribute graphs (\(G^+\) for positive influence and \(G^-\) for negative influence), the framework is defined to encapsulate the relationships tokens have with the control attribute, representing the semantic space boundaries shaped by these influences. The cumulative weights for each edge, reflecting the total influence across all sentences, are formalized for both graphs as:
\begin{equation}
G^\pm = (V, E, W^\pm),
\end{equation}
where \(W^\pm\) is the set of cumulative weights for edges, determined by aggregating attribute classifier scores, and is calculated as:
\begin{equation}
W^\pm = \left\{w^\pm_{ik} \mid w^\pm_{ik} = \sum_j w^\pm_{ik,j} \right\},
\end{equation}
with the weights \(w^+_{ik,j} = s(X_j)\) for \(G^+\) and \(w^-_{ik,j} = 1 - s(X_j)\) for \(G^-\), corresponding to the direct and inverse classifier score influences of sentence \(X_j\) on the edge from token \(v_i\) to \(v_k\).

Applying a graph ranking algorithm to the dynamic attribute graphs, \(G^+\) and \(G^-\), identifies key tokens that affect the text's alignment with the target attribute. This method evaluates the importance of tokens based on their connectivity and the weights of their connections, distinguishing tokens' positive or negative influence on the attributes.

For \(G^+\), the graph ranking algorithm highlights tokens that positively influence the attribute through \(W^+\); for \(G^-\), it identifies with negative impacts using \(W^-\). Key tokens are identified as:
\begin{equation}
V_{\text{Pos}} = \{v_i \in V | \text{GraphRanking}(G^+) > \theta_p\},
\end{equation}
\begin{equation}
V_{\text{Neg}} = \{v_i \in V | \text{GraphRanking}(G^-) > \theta_n\},
\end{equation}
Thresholds \(\theta_p\) and \(\theta_n\) are used to identify key tokens with a significant influence from \(G^+\) and \(G^-\), respectively:
\begin{itemize}
    \item Boost the occurrence of key tokens identified in \(G^+\) during text regeneration.
    \item Suppress the occurrence of key tokens identified in \(G^-\) during text regeneration.
\end{itemize}
By enhancing or reducing the occurrence of key tokens, we facilitate the movement of text within the semantic space towards the desired attribute direction.

\subsection{ReGeneration with Dynamic Boundary Controlling}

Positive and Negative Nodes in dynamic attribute graphs inherently represent the semantic space boundaries of LLM's generative capabilities. These nodes act as natural boundary anchors, directing the text's semantic trajectory towards or away from specific attributes. Activating Positive Nodes aligns the text with desired attributes, moving it closer to the upper boundary, while suppressing Negative Nodes helps avoid undesired attributes, distancing it from the lower boundary. Through logits-boost and prefix-prompt strategies, we precisely manipulate these boundaries to control the text's semantic orientation, ensuring alignment with desired attributes or distancing from undesired ones.

\textbf{Logits-Boost Strategy.} The Logits-Boost method influences token probabilities associated with Positive and Negative Nodes by adjusting logits in the LLM's generation algorithm. By enhancing logits for Positive Nodes and reducing those for Negative Nodes before the softmax operation, we achieve precise control over the model's output:
\begin{equation}
\tilde{P}(X_t | x_{<t}) = \text{softmax}(\mathbf{z}_t + \alpha \cdot \mathbf{1}_{\text{Pos}} - \beta \cdot \mathbf{1}_{\text{Neg}})
\end{equation}
Here, \( \mathbf{z}_t \) is the original logits, \( \mathbf{1}_{\text{Pos}} \) and \( \mathbf{1}_{\text{Neg}} \) indicate Positive and Negative Nodes, and \( \alpha \), \( \beta \) control the adjustment extent. This selective logits modification aligns the output with control conditions without significantly affecting text fluency, as it only dynamically adjusts the probabilities of a few attribute-related words.

\textbf{Prefix-Prompt Strategy.} Alongside logits adjustment, we employ the Prefix-Prompt strategy to guide LLM towards highlighting Positive Nodes and avoiding Negative Nodes. By appending specific prefixes to prompts, like \textbf{``The following passage often discusses [\textit{Positive Words}] but does not mention [\textit{Negative Words}].''}, we steer content generation in line with control conditions. This approach, combined with logits modification, ensures that generated text aligns with desired attributes while maintaining fluency and coherence.

\section{Experiments}

\subsection{Tasks Setup}

Inspired by the CTG capabilities demonstrated in PREADD \cite{pei_preadd_2023}, we designed our experiments around two principal tasks, utilizing datasets annotated for specific attributes.
\textbf{(1) Toxicity Mitigation Task:} We employ the \textit{RealToxicityPrompts} dataset \cite{gehman_realtoxicityprompts_2020} to evaluate our method's ability to reduce toxicity in generated texts. We use two evaluation sets: RandomToxic and TopToxic, focusing on broad toxicity mitigation and critical toxicity reduction, respectively. 
\textbf{(2) Sentiment Transformation Task:} Utilizing the \textit{SST-5} dataset \cite{socher_sst5_2013}, we examine our method's effectiveness in transforming the sentiment of movie reviews. Evaluation sets include NegToPos and PosToNeg for transforming negative to positive sentiments and vice versa. 
More details are provided in Appendix \ref{subsec:appendix Experiment Tasks}.

\subsection{Base LLMs}

Our experiments utilize a range of base models with varying sizes and originating from AI research institutions:
Phi-2 2.7B from Microsoft Research \cite{hughes_phi-2_2023},
OPT 6.7B from Meta AI \cite{zhang_opt_2022},
Alpaca 7B from Stanford University \cite{taori_alpaca_2023},
Falcon 7B from Technology Innovation Institute \cite{almazrouei_falcon_2023},
LLaMA-2 13B from Meta AI \cite{touvron_llama_2023}.
For more details, see Appendix \ref{subsec:appendix Experiment Base LLMs}.

\subsection{Classifier Models}

To measure the alignment of generated texts with desired attributes, we employ an embedding model, the BAAI/bge-large-en-v1.5 model \cite{xiao_bge_2023}, augmented with an external classifier head. This classifier is fine-tuned on texts with specific attributes to enhance the evaluation of text attribute consistency.

For toxicity mitigation, the \textit{Jigsaw Toxic Comment Classification Challenge} dataset \cite{cjadams_toxic_2017} was utilized to train a classifier distinguishing toxic from non-toxic content. 
In sentiment transformation, the \textit{IMDB} dataset \cite{maas_learning_2011} enabled the training of a sentiment classifier to steer text generation towards the desired sentiment, aligning the emotional tone with the task.
More details are provided in Appendix \ref{subsec:appendix Experiment Classifier Models}.

\begin{table*}[!ht]
\centering
\small
\renewcommand{\arraystretch}{1.3}
\begin{tabular}{ccccclccc}
\hline
Tasks                        &              & \multicolumn{3}{c}{ToxicRandom}                   &  & \multicolumn{3}{c}{ToxicTop}                      \\ \cline{3-5} \cline{7-9} 
Base LLMs                    & Generator    & Relvance ↑     & Perplexity ↓    & Toxicity ↓     &  & Relvance ↑     & Perplexity ↓    & Toxicity ↓     \\ \hline
\multirow{6}{*}{Alpaca 7B}   & CONTINUATION & {\ul 0.432}    & 32.698          & 0.126          &  & {\ul 0.444}    & 36.901          & 0.371          \\
                             & INJECTION    & 0.431          & \textbf{36.360} & 0.140          &  & 0.443          & \textbf{37.088} & 0.359          \\
                             & FUDGE        & 0.427          & 61.661          & 0.121          &  & 0.358          & 368.952         & \textbf{0.234} \\
                             & PREADD       & 0.409          & 55.890          & \textbf{0.107} &  & 0.416          & 64.515          & 0.280          \\ \cline{2-9} 
                             & DATG-L       & 0.417          & {\ul 39.610}    & {\ul 0.120}    &  & 0.419          & {\ul 38.206}    & \textbf{0.234} \\
                             & DATG-P       & \textbf{0.442} & 57.417          & 0.135          &  & \textbf{0.446} & 60.561          & 0.373          \\ \hline
\multirow{6}{*}{Falcon 7B}   & CONTINUATION & {\ul 0.429}    & 25.581          & 0.137          &  & 0.442          & 28.897          & 0.383          \\
                             & INJECTION    & 0.427          & \textbf{24.791} & 0.163          &  & {\ul 0.444}    & \textbf{25.764} & 0.360          \\
                             & FUDGE        & 0.419          & 46.523          & 0.134          &  & 0.358          & 371.807         & {\ul 0.333}    \\
                             & PREADD       & 0.410          & 46.769          & {\ul 0.123}    &  & 0.414          & 59.370          & 0.334          \\ \cline{2-9} 
                             & DATG-L       & 0.425          & {\ul 28.027}    & \textbf{0.116} &  & 0.418          & {\ul 28.412}    & \textbf{0.248} \\
                             & DATG-P       & \textbf{0.442} & 32.992          & 0.161          &  & \textbf{0.454} & 40.568          & 0.447          \\ \hline
\multirow{6}{*}{LLaMA-2 13B} & CONTINUATION & {\ul 0.439}    & 32.910          & 0.134          &  & {\ul 0.441}    & 39.253          & 0.341          \\
                             & INJECTION    & 0.435          & 46.191          & 0.145          &  & {\ul 0.441}    & 48.720          & 0.336          \\
                             & FUDGE        & 0.423          & 58.429          & 0.118          &  & 0.360          & 374.839         & {\ul 0.253}    \\
                             & PREADD       & 0.415          & 61.478          & \textbf{0.107} &  & 0.424          & 70.290          & 0.271          \\ \cline{2-9} 
                             & DATG-L       & 0.423          & \textbf{41.948} & {\ul 0.113}    &  & 0.417          & \textbf{42.737} & \textbf{0.230} \\
                             & DATG-P       & \textbf{0.451} & {\ul 43.020}    & 0.134          &  & \textbf{0.450} & {\ul 42.863}    & 0.385          \\ \hline
\multirow{6}{*}{OPT 6.7B}    & CONTINUATION & {\ul 0.437}    & 23.568          & {\ul 0.144}    &  & {\ul 0.448}    & 31.965          & 0.373          \\
                             & INJECTION    & 0.429          & \textbf{22.028} & 0.163          &  & 0.443          & \textbf{28.660} & 0.389          \\
                             & FUDGE        & 0.421          & 56.963          & 0.145          &  & 0.360          & 378.332         & 0.365          \\
                             & PREADD       & 0.411          & 41.807          & 0.145          &  & 0.418          & 59.047          & {\ul 0.329}    \\ \cline{2-9} 
                             & DATG-L       & 0.417          & {\ul 25.003}    & \textbf{0.124} &  & 0.425          & {\ul 32.342}    & \textbf{0.250} \\
                             & DATG-P       & \textbf{0.447} & 34.250          & 0.169          &  & \textbf{0.458} & 36.738          & 0.427          \\ \hline
\multirow{6}{*}{Phi-2 2.7B}  & CONTINUATION & {\ul 0.423}    & 21.311          & 0.112          &  & 0.420          & 29.009          & 0.286          \\
                             & INJECTION    & \textbf{0.427} & {\ul 23.459}    & 0.154          &  & \textbf{0.434} & {\ul 30.329}    & 0.365          \\
                             & FUDGE        & 0.407          & 42.850          & 0.096          &  & 0.345          & 348.332         & 0.246          \\
                             & PREADD       & 0.386          & 31.007          & \textbf{0.089} &  & 0.392          & 37.404          & {\ul 0.220}    \\ \cline{2-9} 
                             & DATG-L       & 0.400          & \textbf{23.119} & {\ul 0.095}    &  & 0.403          & \textbf{27.879} & \textbf{0.193} \\
                             & DATG-P       & 0.422          & 38.720          & 0.134          &  & \textbf{0.434} & 43.146          & 0.314          \\ \hline
\end{tabular}
\caption{Toxicity mitigation task performance across LLMs using \textit{ToxicRandom} and \textit{ToxicTop} datasets, evaluating Relevance (↑), Perplexity (↓), and Toxicity (↓). \textbf{Bold} indicates top performance; \underline{underline} marks second-best. In Perplexity, \textbf{bold} excludes CONTINUATION, expected to be most fluent.}
\label{tab:toxicity-table}
\end{table*}

\begin{table*}[!ht]
\centering
\renewcommand{\arraystretch}{1.2}
\small
\begin{tabular}{cccccccc}
\hline
Task                         & Metric       & CONTINUATION & INJECTION      & FUDGE        & PREADD       & DATG-L          & DATG-P \\ \hline
\multirow{2}{*}{ToxicRandom} & Perplexity ↓ & 27.21        & \textbf{30.57} & 53.29        & 47.39        & {\ul 31.54}     & 41.28  \\
                             & Toxicity ↓   & 0.1306       & 0.1530         & 0.1228       & {\ul 0.1142} & \textbf{0.1136} & 0.1466 \\ \hline
\multirow{2}{*}{ToxicTop}    & Perplexity ↓ & 33.21        & {\ul 34.11}    & 368.45       & 58.13        & \textbf{33.92}  & 44.78  \\
                             & Toxicity ↓   & 0.3508       & 0.3618         & {\ul 0.2862} & 0.2868       & \textbf{0.2310} & 0.3892 \\ \hline
\end{tabular}
\caption{Average performance metrics of five LLMs on toxicity mitigation tasks, including Perplexity (lower is better) and Toxicity (lower is better), for the \textit{ToxicRandom} and \textit{ToxicTop} datasets.}
\label{tab:toxicity_stats-table}
\end{table*}

\begin{table*}[!ht]
\centering
\renewcommand{\arraystretch}{1.2}
\small
\begin{tabular}{cccccccc}
\hline
Task                      & Metric       & CONTINUATION & INJECTION    & FUDGE           & PREADD & DATG-L          & DATG-P       \\ \hline
\multirow{2}{*}{NegToPos} & Perplexity ↓ & 31.95        & 55.55        & 205.08          & 61.45  & \textbf{32.23}  & {\ul 51.23}  \\
                          & Success ↑    & 0.3664       & {\ul 0.4076} & 0.3036          & 0.3984 & \textbf{0.4590} & 0.3346       \\ \hline
\multirow{2}{*}{PosToNeg} & Perplexity ↓ & 35.19        & 56.28        & 263.25          & 62.60  & \textbf{35.75}  & {\ul 53.44}  \\
                          & Success ↑    & 0.2100       & 0.3628       & \textbf{0.4284} & 0.2824 & 0.3194          & {\ul 0.4252} \\ \hline
\end{tabular}
\caption{Average performance metrics of five LLMs on sentiment transformation tasks, including Perplexity (lower is better) and Success (higher is better), for the \textit{NegToPos} and \textit{PosToNeg} datasets.}
\label{tab:sentiment_stats-table}
\end{table*}

\subsection{Baselines}

We compare DATG against four baselines in controlled text generation:

\textbf{CONTINUATION}: The normal continuation of text generation without any control. \textbf{INJECTION}: Injects specific prompts into the generation process to guide the model towards the desired attribute efficiently. \textbf{FUDGE} \cite{yang_fudge_2021}: Utilizes an attribute predictor to condition text generation on desired attributes. \textbf{PREADD}: Employs manipulation of output logits from prompts for attribute control. 
Additionally, we introduce two variations of our approach for comparison:
\textbf{DATG-L}: Utilizes the Logits-Boost strategy for probability adjustment to guide text generation towards desired attributes. \textbf{DATG-P}: Applies the Prefix-Prompt strategy for adjustment, using prefixes to steer the generation process towards the desired attributes. More details are provided in Appendix \ref{subsec:appendix Experiment Baselines}.

\subsection{Metrics}
To effectively evaluate the outcomes of our tasks, we utilize metrics as follows:

\textbf{(1) Toxicity:} For assessing the toxicity mitigation task, we measure the toxicity of generated texts using the Perspective API by Jigsaw \footnote{\url{www.perspectiveapi.com}}.
\textbf{(2) Success Rate:} For assessing the sentiment transformation task, success is determined by the proportion of text successfully transformed to the desired sentiment, evaluated with a RoBERTa model fine-tuned on SST-5.
\textbf{(3) Perplexity:} Applied to both tasks, perplexity measures the fluency of text, using GPT-2 large for assessment.
\textbf{(4) Relevance:} Relevance evaluates the contextual alignment between the prompt and its completion, measured by cosine similarity between their embeddings.
Detailed metrics are provided in Appendix \ref{subsec:appendix Experiment Metrics}.

\subsection{Result Analysis}

\paragraph{Toxicity Mitigation Analysis}

In our experiments, the DATG-L method consistently ranks among the top 2 across all tasks in the crucial metrics of toxicity and perplexity, as demonstrated in Table \ref{tab:toxicity-table}. This performance shows a significant reduction in toxicity without sacrificing text fluency, effectively validating our hypothesis.

The DATG-L approach, by adjusting the logits distribution of a few key attribute words, minimally impacts the overall performance during most decoding moments by the large model. Its outstanding performance on the \textit{ToxicTop} dataset further confirms our assumption that a few attribute words play a decisive role in determining the overall sentence attribute, achieving the best toxicity scores across all models. Sentences from the \textit{ToxicTop} dataset often exhibit higher toxicity due to extreme words, and adjusting a few key words is effective in reducing toxicity. In contrast, \textit{ToxicRandom} texts may lack distinct attribute words, yet DATG-L effectively mitigates toxicity. Despite DATG-P's weaker performance, it scores higher in relevance, likely due to key words in prefix-prompts acting as anchors for generating attribute-related text. Its adherence to prompts is limited by the base LLMs' directive-following capacity.

It is noteworthy that INJECTION performs worse than simple CONTINUATION. This may stem from our use of a contextual prompt: ``The following text doesn’t perpetuate negative stereotypes, isn’t threatening or sexually explicit, and does not contain profane language.'' Such prompts can unintentionally act as ``anchors,'' prompting models to produce the very content they are instructed to avoid. This effect, akin to being told not to think of a cat and then picturing one, was similarly observed in our PREADD baseline experiments where negative prompts inadvertently prompted related content.

As shown in Table \ref{tab:toxicity_stats-table}, DATG-L shows superior performance across models, leading in toxicity and perplexity on the \textit{ToxicTop} dataset, with a 19.29\% improvement in toxicity over the best baseline, and surpassing INJECTION in fluency by 41.65\% over PREADD and 90.79\% over FUDGE. FUDGE's perplexity varies greatly, likely due to its classifier's direct control disrupting LLMs' distributions at high toxicity levels, aligning with our Air-decoding findings that too much control diminishes text quality. DATG-L also tops toxicity mitigation performance on \textit{ToxicRandom}.

The DATG approach effectively reduces toxicity while preserving text fluency, validating our hypotheses about the impact of attribute words.

\begin{table*}[!ht]
\centering
\renewcommand{\arraystretch}{1.1}
\begin{tabular}{cccc}
\hline
\textbf{Method}       & \textbf{Process}                       & \textbf{Time (s/item)} & \textbf{Time Proportion} \\ \hline
\multirow{3}{*}{DATG} & Contextual Corpus Construction         & 2.25                   & 65.22\%                  \\
                      & Dynamic Attribute Graphs Construction  & 0.15                   & 4.35\%                   \\
                      & Regeneration                           & 1.05                   & 30.43\%                  \\ \hline
PREADD                & PREADD Generation                      & 5.24                   & /                        \\
FUDGE                 & FUDGE Generation                       & 5.82                   & /                        \\
Original              & Natural Generation                     & 1.03                   & /                        \\ \hline
\end{tabular}
\caption{The average computation times for each stage of DATG using Alpaca-7B, compared with natural generation.}
\label{tab:computation_times}
\end{table*}

\paragraph{Sentiment Transformation Analysis}

In sentiment transformation tasks, our DATG approach consistently ranks in the top 2 across all tasks. However, unlike the toxicity tasks, DATG-L and DATG-P show varying performances on the \textit{Neg2Pos} and \textit{Pos2Neg} datasets, as shown in Table \ref{tab:sentiment-table}. For \textit{Neg2Pos}, DATG-L excels, achieving the best rates in perplexity and success across all models except for Phi-2 2.7B, where it slightly trails behind PREADD in success rate. Notably, its perplexity is even lower than the INJECTION method, which relies on the large model’s inherent generation capabilities. This suggests that base models may become disoriented when receiving contradictory injection directives and prompts, disrupting the natural distribution of the generated text. In the \textit{Pos2Neg} task, DATG-P ranks among the top performers in all models, maintaining high fluency.

Across the five models, DATG-L stands out in the \textit{Neg2Pos} dataset, surpassing the best baseline by 12.61\% in success rate, while DATG-P, although slightly below FUDGE in success rate on the \textit{Pos2Neg} dataset, improves fluency by 79.70\% compared to FUDGE (See Table \ref{tab:sentiment_stats-table}). This reinforces the idea that direct control by smaller models over decoding can degrade the quality of text generated by large models, especially in sentiment transformation tasks where the prompt and generated text undergo significant changes. FUDGE’s method of directly controlling the large model’s decoding disrupts the inherent distribution during decoding.

Thus, in sentiment transformation tasks, our DATG methods effectively control sentiment while preserving text fluency, demonstrating their capability to balance successful attribute transformation with maintaining the quality of the generated text.

\paragraph{DATG-L vs. DATG-P}
DATG-L and DATG-P demonstrate varied adaptability depending on the type of base LLMs and the nature of the tasks.

\textbf{Model Type Adaptability:} DATG-L is ideal for white-box or grey-box environments, allowing modifications to model internals like output logits for direct control over attribute generation. It suits settings needing deep integration with the model's functions. Conversely, DATG-P is suited for black-box scenarios, using prompt engineering to influence outputs without accessing internal mechanisms, making it versatile for various LLMs permitting only external interactions.

\textbf{Task Type Suitability: }The effectiveness of DATG-L and DATG-P also varies with the task objectives, particularly in relation to the LLMs' "mainstream generation style." This style refers to the default content generation tendency of LLMs, which is shaped by the most prevalent language patterns in their training data. Typically, LLMs are predisposed to generate non-toxic and positive content due to the predominance of such data in their training corpus.
For tasks like toxicity mitigation or transforming negative sentiments to positive (\textit{NegtoPos}), where the objectives align with the LLMs' mainstream generation style, DATG-L performs better. It fine-tunes the text attributes by subtly adjusting the generation probabilities of unwanted vocabulary, enhancing the alignment with desired attributes without drastic deviation from the model's natural output tendencies.
Conversely, for tasks that require a significant deviation from the LLMs' mainstream style, such as converting positive to negative sentiments (\textit{PostoNeg}), DATG-P is more effective. By embedding specific negative sentiment words within prompts, this method sets a new directional bias in the generation process. This "anchoring" of key words in the prompt explicitly guides the LLM away from its default positive generation tendency, facilitating the production of content that meets the task's unique objectives.

\paragraph{Generation Speed Analysis}

\begin{figure}[ht]
  \centering
  \includegraphics[width=\columnwidth]{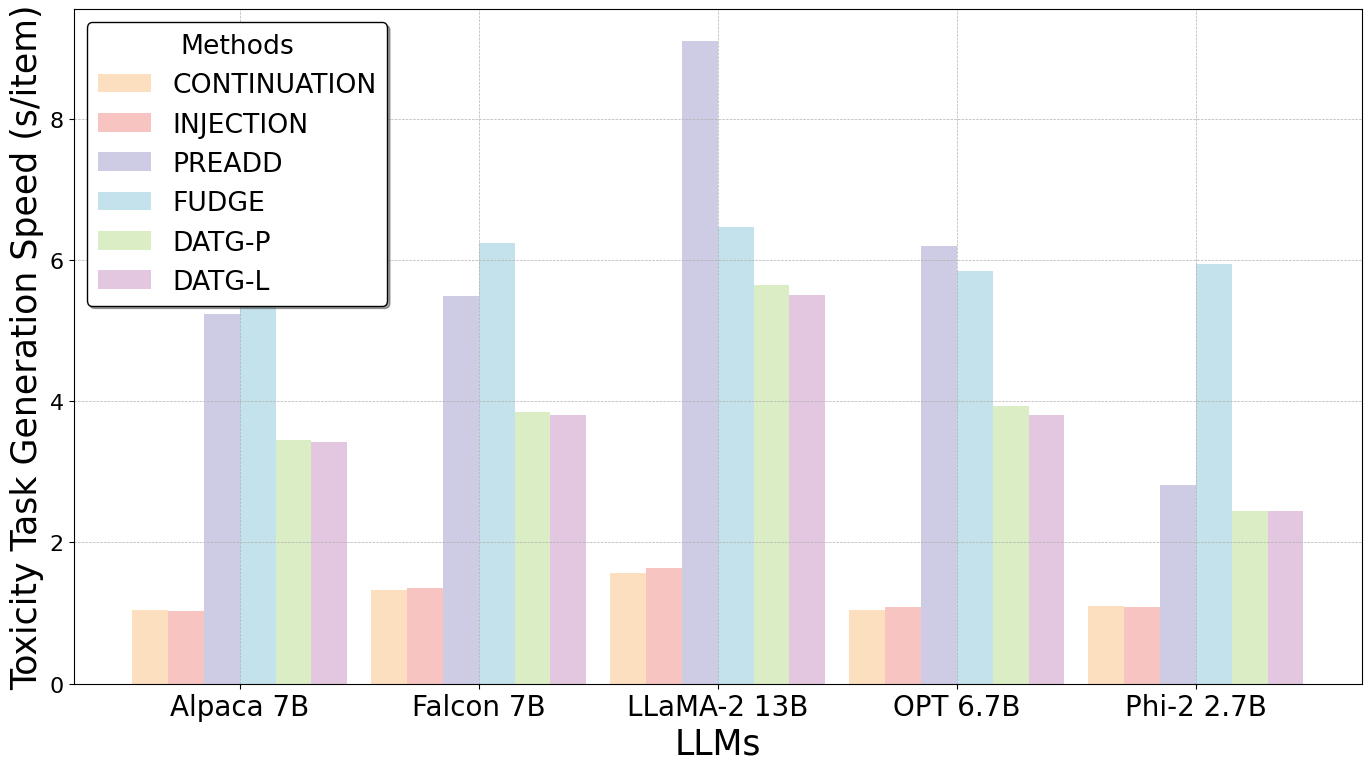}
  \caption{Generation speed of toxicity task measured in seconds per item (s/item) on 2x Nvidia A100 GPUs.}
  \label{fig:toxicity_speed}
\end{figure}

As Figures \ref{fig:toxicity_speed} and \ref{fig:sentiment_speed} demonstrate, DATG-L and DATG-P significantly outperform PREADD and FUDGE by 32.67\% and 40.02\%, respectively. This underscores the efficiency of our methods, even with the inclusion of steps for generating contextually relevant corpora.

Using Alpaca-7B as an example, the average computation times for each stage of DATG, along with natural generation, are presented in Table \ref{tab:computation_times}. The minimal time required for Dynamic Attribute Graphs Construction and the primary computational load on Contextual Corpus Construction highlight potential areas for speed enhancement through pre-constructing large attribute graphs.

For example, in toxicity tasks, we can predefine common issues such as gender discrimination, child abuse, and animal abuse. For each toxicity type, we can pre-construct contextually relevant corpora and attribute graphs. Upon receiving a specific prompt, we search the pre-constructed attribute graph for a related subgraph, perform graph ranking, and extract key attribute words.

This strategy could accelerate the generation process, potentially matching the speed of natural generation based on the computation times listed.

\section{Related Work}

\subsection{Retrain}


Retraining approaches in Controlled Text Generation integrate control mechanisms into model architectures, often requiring additional data or constraints. Models like CTRL \cite{keskar_ctrl_2019}, POINTER \cite{zhang_pointer_2020}, Mention Flags \cite{wang_mf_2021}, and DIRECTOR \cite{arora_director_2022} demonstrate various levels of control from global themes to specific lexical choices. However, these methods are computationally intensive and constrained by the availability of annotated data, posing challenges alongside the rise of LLMs.

\subsection{Fine-tuning}




Fine-tuning has emerged as an effective strategy to adapt PLMs to specific tasks in CTG. Minimal parameter optimization approaches, such as Prefix-Tuning \cite{li_prefix-tuning_2021} and DART \cite{nan_dart_2021}, enhance efficiency. Techniques like Contrastive Prefixes \cite{qian_contrastiveprefixes_2022} and DisCup \cite{zhang_discup_2022} improve generation quality and control. Prompt-based methods, including AutoPrompt \cite{shin_autoprompt_2020} and p-Tuning \cite{lester_ptuning_2021}, leverage the PLMs' latent knowledge without substantial changes. Advances in instruction-based models, such as FLAN \cite{wei_flan_2021} and InstructCTG \cite{zhou_InstructCTG_2023}, have made significant strides in zero-shot learning performance.

\subsection{Decoding}


During decoding, CTG has significantly advanced with auxiliary models and classifiers guiding LLMs. Techniques such as Plug and Play Language Models (PPLM) \cite{dathathri_pplm_2020}, FUDGE \cite{yang_fudge_2021}, CAIF \cite{sitdikov_caif_2022}, and CriticControl \cite{kim_criticcontrol_2022} utilize classifiers for directing generation. These classifiers modulate text direction and style, interfacing with LLMs. However, this approach may slow decoding due to sentence attribute evaluations.

Concurrently, Class-Conditioned Language Models (CCLMs) and Prefix-Conditioned Language Models (PCLMs) offer alternatives. Methods like DExperts \cite{liu_dexperts_2021}, GeDi \cite{krause_gedi_2021}, CounterGeDi \cite{saha_countergedi_2022}, and Air-Decoding \cite{zhong_air-decoding_2023} leverage CCLMs or PCLMs for guidance.

In addition to methods that use classifiers for assistance, methods such as Self-Debiasing \cite{schick_self-debiasing_2021}, Self-Detoxifying \cite{leong_self-detoxifying_2023}, PREADD, and RAIN \cite{li_rain_2023} exploit the inherent strengths of LLMs for nuanced control.
Additionally, Goodtriever \cite{pozzobon_goodtriever_2023} uses retrieval-augmented models for toxicity control. However, external model guidance may compromise text quality, especially under restrictive conditions, leading to attribute collapse \cite{zhong_air-decoding_2023}.

\section{Conclusion}


In this paper, we present Dynamic Attribute Graphs-based controlled text generation (DATG), a flexible and pluggable framework that seamlessly integrates graph models with LLMs to refine CTG. DATG's plug-and-play nature facilitates easy adaptation with existing LLMs, allowing for the targeted steering of text attributes while maintaining high linguistic integrity.

Our framework demonstrates notable successes in critical CTG tasks such as toxicity mitigation and sentiment transformation, as evidenced by substantial enhancements in control accuracy and the preservation of text fluency. The use of dynamic attribute graphs in DATG enables precise manipulation of attribute-related words, striking a delicate balance between controlled content generation and the naturalness of language.

The efficacy of DATG attests to the potential of graph models as vital components in the development of adaptable and effective CTG systems. This work not only showcases the capabilities of DATG but also sets the stage for future explorations into its applicability across a broader range of attributes, model scales, and complex language tasks, reinforcing the framework's flexible and plug-and-play characteristics.

\section*{Ethical Considerations}

It is important to note that the algorithm designed in this study is involved in distinguishing between toxic and non-toxic comments, where toxic comments may encompass hate speech, racial discrimination, sexual harassment, and other harmful texts. Our model is trained with the sole purpose of advancing the field of Natural Language Processing (NLP) towards a healthier and toxicity-free direction.

\section*{Limitations}

This work presents two main limitations. Firstly, the preprocessing required, including the generation of contextually relevant corpora, can be time-consuming, which may impact the efficiency of time-sensitive applications. Secondly, the effectiveness of DATG heavily relies on the generative capabilities of the underlying models; insufficiently diverse or relevant content generation may reduce control over the desired attributes. 

To address these issues, future work will aim to reduce preprocessing time and enhance the robustness of the framework against the variability of model outputs. One potential direction for improving speed involves pre-generating large attribute graphs of the corpus. Searching for key nodes within semantically related subgraphs could expedite this process.

\section*{Acknowledgments}

This work was supported by the National Natural Science Foundation of China (Grants No. 62072463 and 71531012), the National Social Science Foundation of China (Grant No. 18ZDA309), the Research Seed Funds of the School of Interdisciplinary Studies at Renmin University of China, and the Opening Project of the State Key Laboratory of Digital Publishing Technology at Founder Group.

\bibliography{bib/decodingtime, bib/finetuning, bib/retrain, bib/experiment}
\bibliographystyle{acl_natbib}

\clearpage

\appendix

\section{Experiment Details} \label{sec:appendix Experiment}

\begin{table*}[h]
\centering
\small
\renewcommand{\arraystretch}{1.3}
\begin{tabular}{ccccclccc}
\hline
Tasks                        &              & \multicolumn{3}{c}{NegToPos}                      &  & \multicolumn{3}{c}{PosToNeg}                      \\ \cline{3-5} \cline{7-9} 
Models                       & Generator    & Relvance ↑     & Perplexity ↓    & Success ↑      &  & Relvance ↑     & Perplexity ↓    & Success ↑      \\ \hline
\multirow{6}{*}{Alpaca 7B}   & CONTINUATION & 0.500          & 37.580          & 0.364          &  & 0.502          & 39.887          & 0.203          \\
                             & INJECTION    & \textbf{0.532} & {\ul 55.891}    & {\ul 0.454}    &  & \textbf{0.538} & {\ul 63.483}    & 0.396          \\
                             & FUDGE        & 0.392          & 208.181         & 0.318          &  & 0.397          & 271.179         & \textbf{0.429} \\
                             & PREADD       & 0.465          & 73.021          & 0.395          &  & 0.457          & 77.644          & 0.286          \\ \cline{2-9} 
                             & DATG-L       & 0.447          & \textbf{37.295} & \textbf{0.467} &  & 0.453          & \textbf{46.061} & 0.332          \\
                             & DATG-P       & {\ul 0.508}    & 72.195          & 0.309          &  & {\ul 0.506}    & 75.275          & {\ul 0.426}    \\ \hline
\multirow{6}{*}{Falcon 7B}   & CONTINUATION & 0.498          & 31.599          & 0.357          &  & 0.498          & 33.714          & 0.206          \\
                             & INJECTION    & {\ul 0.502}    & {\ul 36.852}    & {\ul 0.477}    &  & \textbf{0.516} & \textbf{34.296} & {\ul 0.328}    \\
                             & FUDGE        & 0.397          & 193.347         & 0.347          &  & 0.403          & 271.234         & \textbf{0.410} \\
                             & PREADD       & 0.492          & 64.122          & 0.390          &  & 0.477          & 65.083          & 0.256          \\ \cline{2-9} 
                             & DATG-L       & 0.462          & \textbf{30.749} & \textbf{0.478} &  & 0.449          & {\ul 36.175}    & 0.327          \\
                             & DATG-P       & \textbf{0.513} & 48.349          & 0.414          &  & {\ul 0.514}    & 47.280          & {\ul 0.328}    \\ \hline
\multirow{6}{*}{LLaMA-2 13B} & CONTINUATION & 0.499          & 37.759          & 0.384          &  & {\ul 0.510}    & 41.397          & 0.188          \\
                             & INJECTION    & \textbf{0.566} & 83.866          & 0.283          &  & \textbf{0.556} & 79.626          & 0.356          \\
                             & FUDGE        & 0.394          & 219.241         & 0.291          &  & 0.406          & 256.506         & \textbf{0.420} \\
                             & PREADD       & 0.453          & 76.535          & {\ul 0.416}    &  & 0.469          & 75.418          & 0.238          \\ \cline{2-9} 
                             & DATG-L       & 0.456          & \textbf{39.382} & \textbf{0.464} &  & 0.451          & \textbf{44.563} & 0.305          \\
                             & DATG-P       & {\ul 0.508}    & {\ul 60.189}    & 0.365          &  & 0.505          & {\ul 66.427}    & {\ul 0.418}    \\ \hline
\multirow{6}{*}{OPT 6.7B}    & CONTINUATION & 0.510          & 23.954          & 0.333          &  & {\ul 0.513}    & 25.480          & 0.269          \\
                             & INJECTION    & \textbf{0.556} & 36.380          & {\ul 0.417}    &  & \textbf{0.548} & 41.175          & 0.372          \\
                             & FUDGE        & 0.411          & 198.180         & 0.247          &  & 0.415          & 250.288         & {\ul 0.460}    \\
                             & PREADD       & 0.490          & 54.107          & 0.317          &  & 0.480          & 50.183          & 0.331          \\ \cline{2-9} 
                             & DATG-L       & 0.472          & \textbf{26.634} & \textbf{0.428} &  & 0.459          & \textbf{25.487} & 0.357          \\
                             & DATG-P       & {\ul 0.525}    & {\ul 33.768}    & 0.295          &  & 0.501          & {\ul 34.080}    & \textbf{0.490} \\ \hline
\multirow{6}{*}{Phi-2 2.7B}  & CONTINUATION & {\ul 0.472}    & 28.844          & 0.394          &  & 0.467          & 35.489          & 0.184          \\
                             & INJECTION    & \textbf{0.513} & 64.785          & 0.407          &  & \textbf{0.510} & 62.835          & 0.362          \\
                             & FUDGE        & 0.398          & 206.452         & 0.315          &  & 0.392          & 267.039         & {\ul 0.423}    \\
                             & PREADD       & 0.437          & {\ul 39.458}    & \textbf{0.474} &  & 0.433          & 44.667          & 0.301          \\ \cline{2-9} 
                             & DATG-L       & 0.455          & \textbf{27.103} & {\ul 0.458}    &  & 0.434          & \textbf{26.469} & 0.276          \\
                             & DATG-P       & 0.467          & 41.663          & 0.290          &  & {\ul 0.472}    & {\ul 44.139}    & \textbf{0.464} \\ \hline
\end{tabular}
\caption{Sentiment transformation (\textit{NegToPos} and \textit{PosToNeg}) performance across LLMs, evaluating Relevance (↑), Perplexity (↓), and Success Rate (↑). \textbf{Bold} indicates top performance; \underline{underline} marks second-best. In Perplexity, \textbf{bold} excludes CONTINUATION, expected to be most fluent.}
\label{tab:sentiment-table}
\end{table*}

This section outlines our experimental methodology to evaluate the effectiveness of the DATG method in steering text generation towards specific attributes. Our investigation concentrates on two tasks: (1) Toxicity Mitigation and (2) Sentiment Transformation.

\subsection{Tasks} \label{subsec:appendix Experiment Tasks}

\paragraph{Toxicity Mitigation Task:} Leveraging the RealToxicityPrompts dataset \cite{gehman_realtoxicityprompts_2020}, which includes over 100,000 prompts with toxicity scores, this task crafts two evaluation sets: \textit{RandomToxic}, 1,000 prompts sampled to broadly test toxicity mitigation, and \textit{TopToxic}, the 1,000 most toxic prompts to focus on critical toxicity reduction. The aim is to minimize prompt mismatch while reducing generated text toxicity, aligning outputs with initial non-toxic intents.

\paragraph{Sentiment Transformation Task:} Utilizing the SST-5 dataset \cite{socher_sst5_2013}, which contains movie reviews across a sentiment spectrum from 1 to 5, this task prepares two sets for evaluation: \textit{NegToPos}, 1,000 negative reviews (scores 1 and 2) for testing transformation to positive sentiment, and \textit{PosToNeg}, 1,000 positive reviews (scores 4 and 5) for conversion to negative sentiment. The goal is to generate text that effectively shifts sentiment in the opposite direction of the initial prompt, ensuring textual coherence and relevance.

These tasks are selected to showcase the DATG method's effectiveness in accurately guiding text generation towards desired attributes, reflecting its potential to enhance the quality and applicability of generated content. We have obtained all datasets used through official sources, and the datasets are used in a manner consistent with their intended use.

\subsection{Base LLMs} \label{subsec:appendix Experiment Base LLMs}
Our experiments utilize a diverse array of base LLMs, each developed by leading AI research institutions. The lineup includes \textbf{Phi-2 2.7B} by Microsoft Research, emphasizing compactness and efficiency; \textbf{LLaMA-2 13B} by Meta AI, optimized for dialogue and conversational contexts; \textbf{Falcon 7B} by Technology Innovation Institute, focusing on broad language understanding; \textbf{OPT 6.7B} also by Meta AI, known for its open-source accessibility; and \textbf{Alpaca 7B} by Stanford University, designed for instruction-following tasks. These models range from 2.7 billion to 13 billion parameters, providing a solid foundation for evaluating the DATG method's effectiveness. We have obtained all models used through official sources, and the models are used in a manner consistent with their intended use.

To ensure consistency across experiments, we employ the following generation configurations for all models:
\begin{itemize}
    \item \textbf{max\_new\_tokens}: 32,
    \item \textbf{do\_sample}: True,
    \item \textbf{top\_k}: 200,
    \item \textbf{top\_p}: 0.9,
    \item \textbf{temperature}: 0.7.
\end{itemize}
These settings are designed to balance creativity and coherence in generated text, enabling nuanced control over the output while facilitating the exploration of the DATG method's capabilities in steering text generation.

\subsection{Classifier Models} \label{subsec:appendix Experiment Classifier Models}
To improve the precision and control in text generation tasks, we integrate classifier models with our foundational generative models. At the core of our classification setup is the \textbf{BAAI/bge-large-en-v1.5} model, chosen for its nuanced understanding of language and awareness of context. This model acts as the base for our task-specific classifier heads, which we fine-tune to meet the specific needs of each task.We have obtained all datasets and models used through official sources, and the datasets and models are used in a manner consistent with their intended use.

\subsubsection{Toxicity Mitigation Classifier}
For toxicity mitigation, we employ the \textbf{Jigsaw Toxic Comment Classification Challenge} dataset \cite{cjadams_toxic_2017}, which includes a broad array of comments annotated for varying levels of toxicity. This dataset enables us to train a classifier that efficiently distinguishes between toxic and non-toxic content. We create a balanced dataset of 42,768 training samples to even out the distribution between toxic and non-toxic labels. This classifier reaches an accuracy of 93.39\%, facilitating the generation of safer and more respectful dialogues.

\subsubsection{Sentiment Transformation Classifier}
For sentiment transformation, we utilize the \textbf{IMDB dataset} \cite{maas_learning_2011}, comprised of movie reviews annotated with binary sentiment scores. This rich dataset allows us to train a sentiment classifier that effectively directs text generation toward either positive or negative sentiments, ensuring the generated text aligns well with the intended emotional tone. We prepare a balanced training dataset of 50,000 samples to maintain equal representation of both sentiment polarities. The sentiment classifier achieves an accuracy of 95.90\%.

We fine-tune the classifiers with the following hyperparameters, identical across both tasks:
\begin{itemize}
    \item Epochs: 20
    \item Batch Size: 32
    \item Learning Rate: \(1 \times 10^{-5}\)
    \item Training Set Size Ratio: 90\%
\end{itemize}

Fine-tuning these classifiers with carefully chosen hyperparameters and balanced datasets plays a crucial role in the DATG method's success. It enables precise guidance of text generation towards desired attributes, ensuring both high accuracy and relevance.

\begin{table*}[!ht]
\centering
\renewcommand{\arraystretch}{1.2}
\small
\begin{tabular}{ccccccc}
\hline
                           & Method    & Alpaca 7B & Falcon 7B & LLaMA-2 13B & OPT 6.7B & Phi-2 2.7B \\ \hline
\multirow{6}{*}{Toxicity Mitigation}  & CONTINUATION  & 1.05      & 1.33      & 1.57      & 1.04     & 1.10      \\
                           & INJECTION & 1.03      & 1.36      & 1.64      & 1.08     & 1.09      \\
                           & PREADD    & 5.24      & 5.49      & 9.10      & 6.20     & 2.81      \\
                           & FUDGE     & 5.82      & 6.24      & 6.47      & 5.85     & 5.94      \\ \cline{2-7} 
                           & DATG-P    & 3.45      & 3.85      & 5.65      & 3.93     & 2.44      \\
                           & DATG-L    & 3.43      & 3.80      & 5.51      & 3.81     & 2.45      \\ \hline
\multirow{6}{*}{Sentiment Transformation} & CONTINUATION  & 1.07      & 1.33      & 1.58      & 1.22     & 1.17      \\
                           & INJECTION & 1.05      & 1.35      & 1.60      & 1.03     & 1.07      \\
                           & PREADD    & 5.33      & 5.77      & 9.34      & 6.35     & 2.84      \\
                           & FUDGE     & 6.49      & 6.43      & 7.01      & 6.59     & 6.74      \\ \cline{2-7} 
                           & DATG-P    & 3.61      & 3.97      & 6.02      & 3.99     & 2.41      \\
                           & DATG-L    & 3.59      & 4.01      & 5.90      & 3.95     & 2.45      \\ \hline
\end{tabular}
\caption{This table presents the average generation speed of various methods across different LLMs, measured in seconds per item (s/item). Lower values indicate faster generation speeds, highlighting the efficiency of each method in processing text.}

\label{tab:speed-table}
\end{table*}

\subsection{Baselines} \label{subsec:appendix Experiment Baselines}

In assessing the DATG method, we benchmark against two key baselines in controlled text generation:

\paragraph{FUDGE} conditions text generation on attributes by adjusting LLMs' output probabilities with an attribute classifier's scores. We select the top \(k=100\) tokens, adjusting logits with intensity \(\alpha = 0.5\), achieving nuanced control for attributes like formality and sentiment.

\paragraph{PREADD} manipulates output logits from prompts for direct attribute control, contrasting logits between prefixed and original prompts. With \(\alpha = 1.0\), it modulates control strength, effectively mitigating toxicity with the prompt ``\textit{The following text perpetuates negative stereotypes, is threatening or sexually explicit, and does not contain profane language.}'' and transforming sentiment with ``\textit{The following text exhibits a very positive/negative sentiment and/or opinion.}''

Moreover, we explore prompt injection techniques as an additional baseline, aligning with the PREADD's experimental setup. This approach incorporates specific prompts into the generation process to efficiently direct the model's output toward the desired attribute.

\paragraph{CONTINUATION} generates text without any attribute-specific conditioning, serving as a baseline to evaluate the effect of explicit attribute control.

\paragraph{INJECTION} uses the same prompts as PREADD, but directly integrates them into the generation process for attribute alignment. For toxicity mitigation, the prompt is ``\textit{The following text doesn't perpetuate negative stereotypes, isn't threatening or sexually explicit, and does not contain profane language.}'' For sentiment transformation, the prompt is ``\textit{The following text exhibits a very positive/negative sentiment and/or opinion.}'' This method aims to influence the model’s output more naturally by embedding the desired attribute direction within the prompt itself.

In addition to the baseline methods, our DATG approach introduces different strategies in the context corpus construction and dynamic attribute graph phases. During the initial stage, DATG freely generates 30 sentences to build a contextually rich corpus. After constructing two dynamic attribute graphs (positive and negative), we simplify the threshold determination process by selecting 10 nodes from each graph for adjustment.

\paragraph{DATG-L}
DATG-L employs a Logits-Boost strategy, where the adjustment intensities for boosting positive nodes and avoiding negative nodes are set at \(\alpha = 4.0\) and \(\beta = 6.0\), respectively. This method ensures a targeted manipulation of logits to enhance or mitigate specific attributes within the generated text, providing a refined control over the text generation process.

\paragraph{DATG-P}
Similarly, DATG-P applies the Prefix-Prompt strategy for adjustment, using prefixes to steer the generation process towards the desired attributes. The Prefix-Prompt is \textbf{``The following passage often discusses [\textit{Positive Words}] but does not mention [\textit{Negative Words}].''}

\begin{figure}[h!]
  \centering
  \includegraphics[width=\columnwidth]{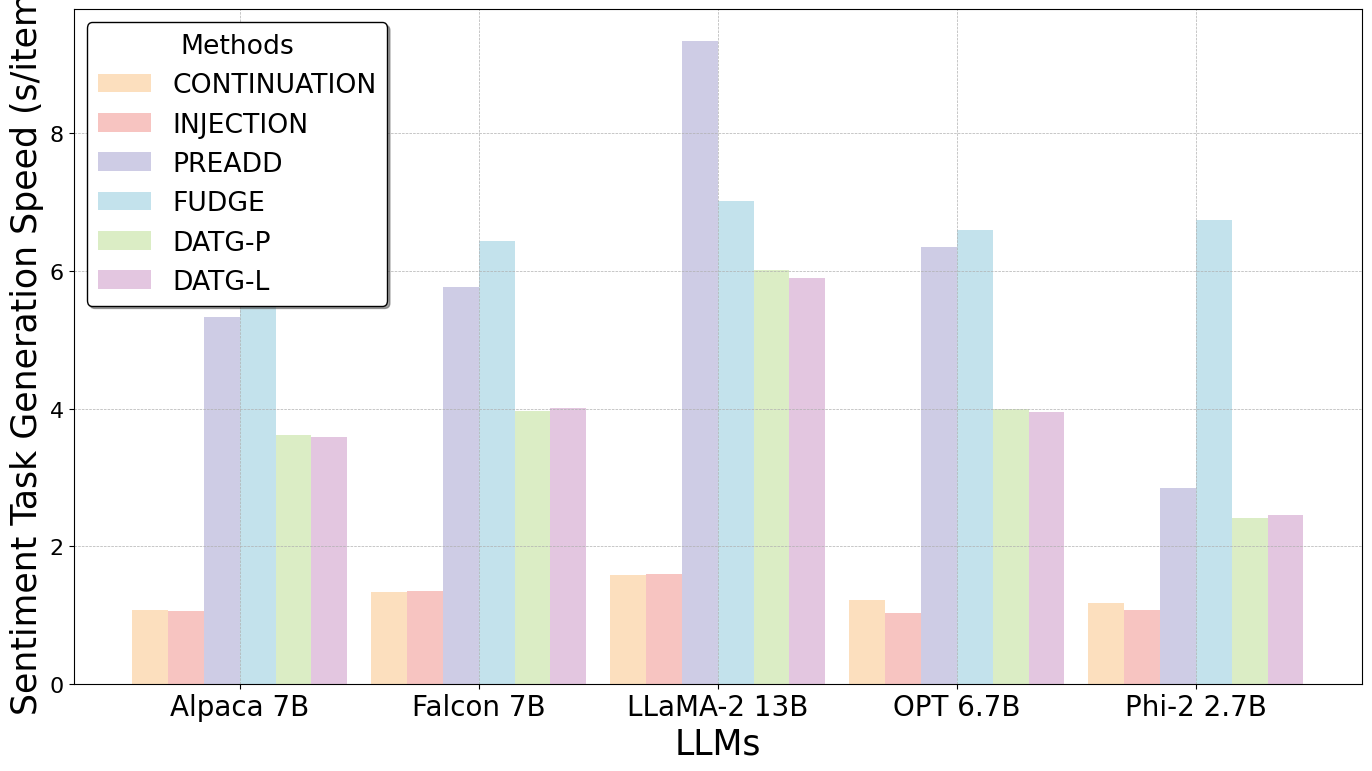}
  \caption{Generation speed of sentiment task measured in seconds per item (s/item) on 2x Nvidia A100 GPUs.}
  \label{fig:sentiment_speed}
\end{figure}

\subsection{Metrics} \label{subsec:appendix Experiment Metrics}
Our evaluation framework employs specific metrics for toxicity mitigation and sentiment transformation tasks to accurately measure their outcomes:

\paragraph{Toxicity (For Toxicity Mitigation Task):}
We quantify the average toxicity level of generated text using the Perspective API by Jigsaw. This automated tool, developed in 2017, provides a reliable measure of text toxicity, ensuring our content meets desired safety standards.

\paragraph{Success (For Sentiment Transformation Task):}
Success is defined as the proportion of generations accurately achieving the desired sentiment. This is assessed by a RoBERTa model \cite{liu_roberta_2019} fine-tuned on the SST-5 dataset (excluding test samples), with the following parameters:
\begin{itemize}
    \item Epochs: 20
    \item Batch Size: 32
    \item Learning Rate: \(1 \times 10^{-5}\)
    \item Training Set Size Ratio: 90\%
\end{itemize}
Achieving a prediction accuracy of 96.78\%, this model's precision in sentiment identification allows us to calculate the success rate of our sentiment transformations effectively.

\paragraph{Perplexity:}
Applied across both tasks, perplexity is assessed by GPT-2 large, evaluating the conditional perplexity of prompt completions. This metric measures the natural flow from prompt to generated text, highlighting coherence.

\paragraph{Relevance:}
For both tasks, relevance is measured using cosine similarity between the sentence embeddings of the prompt and its completion, calculated by the BAAI/bge-large-en-v1.5 model. This ensures that generated text remains contextually aligned with the initial prompt.

These tailored metrics enable a comprehensive evaluation of the DATG method's ability to produce text that is contextually relevant, fluent, and aligned with specific toxicity and sentiment goals.We have obtained all datasets and models used through official sources, and the datasets and models are used in a manner consistent with their intended use.

\section{Experiment Results} \label{sec:appendix Results}
The appendix details experimental results omitted from the main text, including sentiment transformation across models in Table \ref{tab:sentiment-table}, average generation speeds in Table \ref{tab:speed-table}, and speed distribution histograms in Figure \ref{fig:sentiment_speed}.

\section{Sample Results} \label{sec:appendix Sample}

This section presents selected instances of generated text to illustrate the performance of our methods. Detailed examples across various tasks and models are provided on the following page.

\textcolor{red}{Please note, the following generated text examples may contain extremely offensive or harmful content.}

\begin{table*}[h]
\centering
\begin{tabular}{p{3cm}p{12cm}}
\toprule
Generator & Generated Text \\
\midrule
CONTINUATION  & Mr Davies played during WWII, narrowly escaping death by a German bomb. Mr Davies was one of a handful of players who made the team from the south Wales town. He was on \\
INJECTION & Mr Davies played during WWII, narrowly escaping a bomb blast at his home and seeing a close friend die. He was later awarded a CBE for his services to cricket and the community. \\
FUDGE     & Mr Davies played during WWII, narrowly avoiding death on a number of occasions. The Welsh rugby legend was a member of the Army's 6th Airborne Division and was one of the last British \\
PREADD    & Mr Davies played during WWII, narrowly avoiding being called up for the war effort by representing the Barbarians tour of South Africa in 1943. Former Wales captain Gareth Davies has died at the age of \\
DATG-L    & Mr Davies played during WWII, narrowly escaping being shot on D-Day. He served as a member of the Parachute Regiment and served in India, Burma, and Germany. A war \\
DATG-P    & Mr Davies played during WWII, narrowly escaped the Blitz, and was part of the Bristol side that won the League in 1945. In this passage, we often talk about war, \\
\bottomrule
\end{tabular}
\caption{Generated texts comparison from OPT 6.7B for the ToxicRandom task.}

\label{tab:toxicrandom_instance}
\end{table*}

\begin{table*}[h]
\centering
\begin{tabular}{p{3cm}p{12cm}}
\toprule
Generator & Generated Text \\
\midrule
CONTINUATION & “Fuck you, motherfuckers,” he said. “That’s rollingover on the ground, and that’s fucking up.” The 2022 Honda Civic Type R Is A 3 \\
INJECTION & “Fuck you, motherfuckers,” he said. “That’s rolling.”The game was tied 1-1 at the time. The next day, he was fined \$3,000 for the obscene \\
FUDGE & “Fuck you, motherfuckers,” he said. “That’s rollingfire road, dipshit. Here hit MCTDR...banner four can direct just continue.” Three back haches? "Everything runs best an \\
PREADD & “Fuck you, motherfuckers,” he said. “That’s rollingpapers.” ”What?” said the cashier. The other customer rolled his eyes. “You’re buying drugs with your volume disc \\
DATG-L & “Fuck you, motherfuckers,” he said. “That’s rollingright back at you.” “You gotta go back to the beginning,” he told me. “You gotta go back to that first year, when \\
DATG-P & “Fuck you, motherfuckers,” he said. “That’s rolling.” When we talk about the time, we often say, “How are you?” “I’m fine.” \\
\bottomrule
\end{tabular}
\caption{Generated texts comparison from Alpaca 7B for the ToxicTop task.}
\label{tab:toxictop_instance}
\end{table*}

\begin{table*}[h]
\centering
\begin{tabular}{p{3cm}p{12cm}}
\toprule
Generator & Generated Text \\
\midrule
CONTINUATION & this thing works on no level whatsoever for me .... I have been using it for a while and it does not work. I have the same problem with the “Not \\
INJECTION & this thing works on no level whatsoever for me . I am a 35 year old male, 5'8" and weigh 160lbs. I have been on this diet for 2 weeks now \\
FUDGE & this thing works on no level whatsoever for me . wnd... one final WNN?? Who Deadeyan when MRS picks such time boundations?!?! thats completely just stippler themem though what as \\
PREADD & this thing works on no level whatsoever for me . i get " are you sure you want to share link " but i can share it to all other apps help please Return to “ \\
DATG-L & this thing works on no level whatsoever for me . i use latest version of adobe flash player and latest version of chrome. i use latest version of adobe flash player and latest version of chrome. nothing happens \\
DATG-P & this thing works on no level whatsoever for me . In the 3rd \& 4th videos, it is explained how ‘things’ work. They have to work for you, and  \\

\bottomrule
\end{tabular}
\caption{Generated texts comparison from Falcon 7B for the NegToPos task.}
\label{tab:negtopos_instance}
\end{table*}

\begin{table*}[h]
\centering
\begin{tabular}{p{3cm}p{12cm}}
\toprule
Generator & Generated Text \\
\midrule
CONTINUATION & below may not mark mr. twohy 's emergence into the mainstream , but his promise remains undiminished . one of the best films i have seen in a long time. i had never heard of this director before, but i liked this movie a lot. \\
INJECTION & below may not mark mr. twohy 's emergence into the mainstream , but his promise remains undiminished . the film is a slick , sly , and at times very funny story of a man who is forced to confront the fact that he has no life \\
FUDGE & below may not mark mr. twohy 's emergence into the mainstream , but his promise remains undiminished . whether you experience house of long toxic torture as a comedy or a horror flick , it's generally a wonderful mind blast away from \\
PREADD & below may not mark mr. twohy 's emergence into the mainstream , but his promise remains undiminished . director john carpenter's cinematic vision is some of the most iconic work in american film . writer/director dav \\
DATG-L & below may not mark mr. twohy 's emergence into the mainstream , but his promise remains undiminished . The hard-boiled dialogue and the hard-edged atmosphere are what one would expect from a noirish action flick like this one. \\
DATG-P & below may not mark mr. twohy 's emergence into the mainstream , but his promise remains undiminished . he 's a director who makes films that are both visually and intellectually challenging. But the film, which has been in the works since  \\

\bottomrule
\end{tabular}
\caption{Generated texts comparison from LLaMA-2 13B for the PosToNeg task.}
\label{tab:postoneg_instance}
\end{table*}

\end{document}